\documentclass[sigconf]{acmart}

\usepackage[inkscapeformat=png]{svg}

\usepackage{tcolorbox}
\usepackage{listings}

\AtBeginDocument{%
  \providecommand\BibTeX{{%
    \normalfont B\kern-0.5em{\scshape i\kern-0.25em b}\kern-0.8em\TeX}}}

\setcopyright{acmlicensed}
\copyrightyear{2018}
\acmYear{2018}
\acmDOI{XXXXXXX.XXXXXXX}

\acmConference[Conference acronym 'XX]{Make sure to enter the correct
  conference title from your rights confirmation emai}{June 03--05,
  2018}{Woodstock, NY}
\acmISBN{978-1-4503-XXXX-X/18/06}

\begin{document}

\title{Language-Based User Profiles for Recommendation}

\author{Yijia Dai}
\authornote{Both authors contributed equally to this research.}
\email{yd73@cornell.edu}
\orcid{0009-0006-6431-0851}
\author{Joyce Zhou}
\authornotemark[1]
\email{jz549@cornell.edu}
\orcid{0000-0003-1205-3970}
\affiliation{
  \institution{Cornell University}
  \city{Ithaca, NY}
  \country{USA}}

\author{Thorsten Joachims}
\email{tj36@cornell.edu}
\affiliation{%
  \institution{Cornell University}
  \city{Ithaca, NY}
  \country{USA}}
\orcid{0000-0003-3654-3683}


\begin{abstract}
Most conventional recommendation methods (e.g., matrix factorization) represent user profiles as high-dimensional vectors. 
Unfortunately, these vectors lack interpretability and steerability, and often perform poorly in cold-start settings. 
To address these shortcomings, we explore the use of user profiles that are represented as human-readable text. 
We propose the Language-based Factorization Model (LFM), which is essentially an encoder/decoder model where both the encoder and the decoder are large language models (LLMs). 
The encoder LLM generates a compact natural-language profile of the user's interests from the user's rating history. 
The decoder LLM uses this summary profile to complete predictive downstream tasks. 
We evaluate our LFM approach on the MovieLens dataset, comparing it against matrix factorization and an LLM model that directly predicts from the user's rating history. 
In cold-start settings, we find that our method can have higher accuracy than matrix factorization. 
Furthermore, we find that generating a compact and human-readable summary often performs comparably with or better than direct LLM prediction, while enjoying better interpretability and shorter model input length.
Our results motivate a number of future research directions and potential improvements.


\end{abstract}




\received{20 February 2007}
\received[revised]{12 March 2009}
\received[accepted]{5 June 2009}

\maketitle

\section{Introduction}

With the current advancements in large language models (LLMs), there is newfound potential for making recommendation platforms more transparent and steerable through the use of natural language \cite{liu2023chatgpt}.
While other works have already explored the use of natural language to explain the predictions of conventional recommendation models \cite{ren2023representation,li2023ctrl,xi2023openworld}, we explore whether natural language itself can be the medium for representing user profiles.
Such language-based user profiles would not only eliminate the need for post-hoc explanations of otherwise unintelligible profiles, but any changes to the language-based profile would causally affect the subsequent recommendations.
This makes language-based profiles naturally intelligible and allows users to steer recommendations by directly initializing or editing their profiles.



\begin{figure*}
    \centering
    \includegraphics[width=\linewidth]{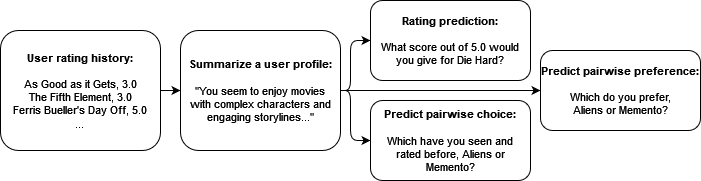}
    \caption{
    Summary of how our user representation method works and what tasks we tested it on
    }
    \label{fig:flowchart}
\end{figure*}

In this paper we propose an encoder/decoder architecture that uses natural language for representing compact user profiles.
In analogy to matrix factorization models we call this architecture a {\em Language-based Factorization Model (LFM)}, and it is illustrated in Figure \ref{fig:flowchart}.
The encoder is an LLM that translates the user rating history into a natural-language profile that summarizes the user's preferences.
The decoder is a separate LLM that takes the natural-language profile and solves various prediction tasks (e.g., rating prediction, pairwise preference prediction, rating validation).
Note that in both the encoder and the decoder stage, the LLMs can take into account natural-language descriptions of the items, providing interesting opportunities for improving cold-start predictions. 


This paper provides a first characterization of how feasible, adaptable, and effective such LFM models can be.
In particular, we focus on evaluating the prediction accuracy that an LFM model can achieve across a range of tasks compared to LLM models that lack the interpretability of a intermediate natural language profile, as well as conventional matrix-factorization methods.
We find that using natural language as a compact profile representation is competitive with an LLM that has no such restriction. In comparison to conventional matrix factorization models, LFM performs competitively in cold start settings, but our zero-shot approach misses the ability to improve when more data is available. We conclude that finetuning both the encoder and the decoder of the LFM provides a promising direction for future work.


    



\section{Related Work}

Many current works \cite{chen2023palr,li2023teach,lyu2023llm} use prompt engineering techniques to enable recommender systems to be personalized.
However, their methods share the same issue with the deep neural networks training as the pipeline is black-box styled and uninterpretable.
Recent works have tried using LLMs on tasks that are traditionally performed by matrix factorization or deep neural network training.
For example, ICAE \cite{ge2023context} essentially trains a LLM \cite{hu2021lora} to generate and use context tokens with LLM prompts for a wide range of tasks, including item recommendation.
However, they do not use the advantage of LLMs to address human-readability for any of these context tokens.
PALR \cite{chen2023palr} and P5/OpenP5 \cite{geng2023recommendation,xu2023openp5} proceed to finetune a LLM based on user interaction history, which has helped the model to outperform other LLM methods on prediction accuracy.
But their method still lacks readability and interpretabiliy, leaving user context largely within the model parameters.
Sanner \cite{sanner2023large} has demonstrated how LLMs can use both a user's liked items or preference descriptions to generate a good set of cold start recommendations.
However, they assume users are manually giving a recommender system some known preferences or a text summary, and have not explored explicitly summarizing the users' previously known preferences to human-readable form.

\section{Methods}


Our evaluation is conducted in a standard recommendation setting, where users have provided cardinal ratings in the past.
We evaluate predictions for new recommendations on the following three tasks, chosen to test model adaptability.
The first is {\em rating prediction}, where the goal is to predict the cardinal rating of a test item.
The second is {\em pairwise preference prediction}, where the model predicts which one of two test items is rated higher by the user.
And the third is {\em pairwise choice prediction}, where the model predicts which one of two test items the user has chosen to watch and rate before.

For each of these tasks, we compare the following four methods.
All methods use a subset of user rating history as training data and input for each user as appropriate.
\begin{description}
    \item[LFM:] This is the proposed language-based factorization model outlined above and illustrated in Figure \ref{fig:flowchart}.
    There is a different decoder for each task, but the encoder generating the natural language profile based on past ratings is shared across all tasks.
    
    \item[LLM-Direct:] This uses a LLM to directly perform prediction tasks based on user rating history without creating any intermediate profile.
    
    \item[NMF:] This is a standard non-negative matrix factorization model\footnote{using \href{https://surprise.readthedocs.io/en/stable/matrix_factorization.html\#surprise.prediction_algorithms.matrix_factorization.NMF}{scikit-surprise unbiased NMF implementation}} \cite{Hug2020}.
    For rating and preference prediction, it is trained on previous rating values.
    For choice prediction, it is trained on previous seen status and uses a supplementary dataset of randomly sampled ``probably unseen'' movie IDs to simulate the unclear boundaries of a user rating history.
\end{description}

We test Llama 2 7B, Llama 2 13B \cite{touvronLlamaOpenFoundation2023}, and Sakura-SOLAR 10.7B\footnote{\url{https://huggingface.co/kyujinpy/Sakura-SOLAR-Instruct}} in instruct (chat) mode downloaded from Hugging Face, using zero-shot prompt tuning.
We follow the standard prompting style used for chat mode for each model.
Hard prompt tuning was done using a validation dataset to improve model accuracy, runtime, and reliability.
Exact prompts and hyperparameters used in our final experiments for all LLM and NMF methods are discussed in appendix \ref{appendix:prompt}.
We also discuss runtime in appendix \ref{appendix:runtime}.

In cases where prediction failed (e.g. matrix factorization encounters an entirely new movie or LLM output is unusable), we impute the mean across all training ratings and substitute that in place of any missing movie ratings.
The \textbf{Default} model is used as a baseline where all rating predictions are substituted with this mean estimate and all pairwise tasks are designed to have accuracy 0.5 (random guessing).

Source code for our experiments is available\footnote{will be on GitHub, not currently set to public as of submission date}.

\section{Dataset and Experiment Setup}
\label{dataset}

We used the MovieLens Tag Genome Dataset 2021 \cite{kotkovRevisitingTagRelevance2021, vigTagGenomeEncoding2012} \footnote{\url{https://grouplens.org/datasets/movielens/tag-genome-2021/}} for the following experiments.
From this large dataset, we randomly sample 300 users. We focus on "typical" users, which we define as users with exactly 150 movie ratings in their history.
This helps us avoid sampling outlier users who have unusual rating patterns (there was a long tail of users with several or several dozen thousand movie ratings, and many users who only have several ratings).

For our experiments, we vary the user history size $c$, using a random subset of size $c$ from the movies the user has rated to generate user representation texts.
For the rating prediction task, for each user-history sample, we collect 3 ``excluded'' movies that users had given ratings for but were not included within their training input.
For each of the pairwise preference prediction and pairwise choice prediction tasks, we similarly collect 3 pairs of ``excluded'' movies based on each history sample where each pair included two movies with different user ratings or different watch choices based on the task type.
This user-history and task ID sampling is repeated 3 times to reduce noise.
This provides 2700 total test points: 3 movies or movie pairs for each user profile task, 3 user profiles based on varying user-history sizes for each MovieLens user, and 300 total MovieLens users.





\section{Experiment Results}

The following describes our key findings.



\paragraph{\bf How often do the methods fail to make a prediction?}

Sometimes the LLM models produce a prediction that does not give any clear answer or is hard for us to parse.
For example, ``I would give it a rating of 4 or 4.5 out of 5'', or ``I am unable to give this a score''.
Similarly, matrix factorization can only guess values for movies that never appeared in training data.
We call these cases ``unreadable'', and the overall parse success rate ``reliability''.


\begin{figure}
    \centering
    \includegraphics[width=\linewidth]{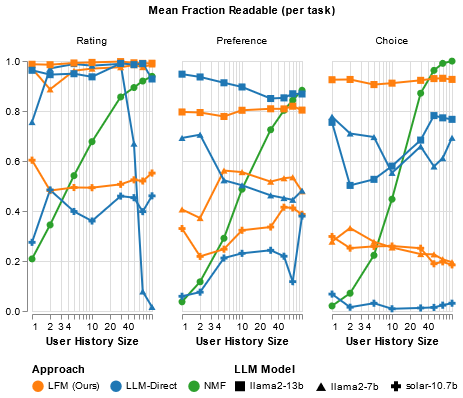}
    \caption{
    Fraction of readable predictions for all tasks with different methods and models vs history size.
    }
    \label{fig:reliable_perhistory}
\end{figure}

In Figure \ref{fig:reliable_perhistory}, we show reliability for different approaches, tasks, and models.
In most cases, our LFM pipeline has comparable or better reliability compared to the appropriate LLM-Direct counterpart.

SOLAR is generally the least reliable in our experiments, although it still demonstrates LFM being more reliable than LLM-Direct.
This could be because SOLAR is trained to explicitly refuse uncertain tasks and may also use different wording compared to the Llama models.
We also note that LLM-Direct with Llama 2 7B becomes significantly less reliable with large history sizes due to exceeding context token limits.

For the following discussion, we focus on task performance with Llama 2 13B, as it demonstrated the highest rates of readable predictions across a range of tasks.
We include task performance results for all other models in Appendix \ref{appendix:experimentsfull}.

\paragraph{\bf How does LFM compare against other methods?}

\begin{figure}
    \centering
    \includegraphics[width=\linewidth]{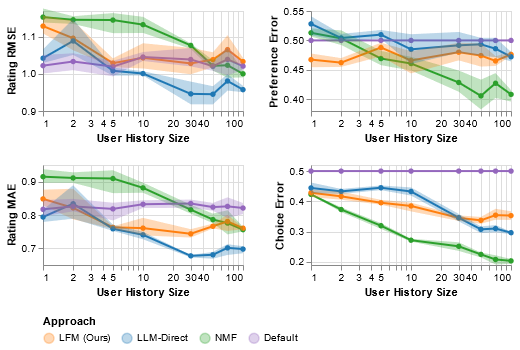}
    \caption{
    Performance (RMSE, MAE, and error rate) for all tasks with different methods (using Llama 2 13B) vs history size.
    }
    \label{fig:alltask_aggrhistory_permodel_bg0_error}
\end{figure}

\begin{figure}
    \centering
    \includegraphics[width=\linewidth]{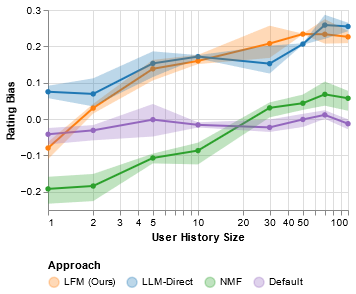}
    \caption{
    Bias (mean error) of rating prediction task with different methods and models vs history sizes.
    }
    \label{fig:scoretask_aggrhistory_permodel_bg0_bias}
\end{figure}

Figure \ref{fig:alltask_aggrhistory_permodel_bg0_error} compares the test error rates for rating prediction (both RMSE and MAE), pairwise preference prediction, and pairwise choice prediction.
For the pairwise tasks, we see no systematic degradation of accuracy from using the profile summary in LFM compared to the LLM-direct approach.
For the rating prediction task, the direct approach shows an advantage for large user-history sizes.
Figure \ref{fig:scoretask_aggrhistory_permodel_bg0_bias} shows that a substantial amount of error for rating prediction is due to bias, which we partly attribute to the fact that LFM and LLM-direct tend to make integer-valued predictions.

Compared to the conventional NMF, the LFM is most competitive in the cold-start regime.
As we will further analyze in later experiments, we conjecture that moving from a zero-shot approach to a fine-tuned version of LFM is a promising direction, so that it could also take advantage of increasing amounts of cross-user data.

\paragraph{\bf How does LFM accuracy vary with the profile size?}

\begin{figure}
    \centering
    \includegraphics[width=\linewidth]{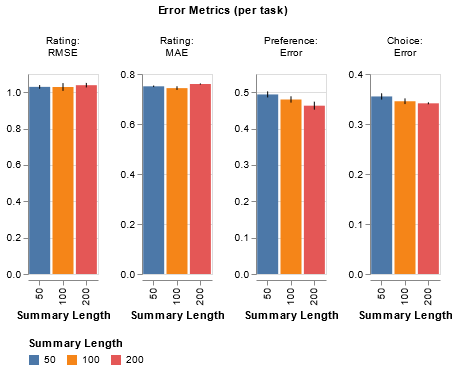}
    \caption{
    Performance metrics (RMSE, MAE, and error rate) for all tasks with different LFM summary lengths with history size 30.
    }
    \label{fig:alltask_30history_permodel_bg0_allsummlength_error}
\end{figure}

Figure \ref{fig:alltask_30history_permodel_bg0_allsummlength_error} shows that varying the profile size in LFM from 50 to 200 words for a user history of 30 items does not have a consistent impact on prediction error. In particular, cardinal rating prediction does not improve with increasing profile size, and the improvements for preference and choice prediction are modest.
Overall, it appears that even a short profile can capture much of the relevant information.
Some example profiles are shown in Appendix \ref{app:summaries}.
 
We note that predicting user choices has the best error rate, and Figure \ref{fig:alltask_aggrhistory_permodel_bg0_error} shows that increases in user history result in continued improvements.
One possible explanation is that short profile texts from LFM largely capture what types of movies the user chooses and rates (e.g., genre, plot), but have less space to suggest specific attributes which go into higher or lower ratings.
Longer profile texts tend to improve pairwise preference error rate, which supports this idea.


\paragraph{\bf How does LFM compare against NMF with varying amounts of background data?}

\begin{figure}
    \centering
    \includegraphics[width=\linewidth]{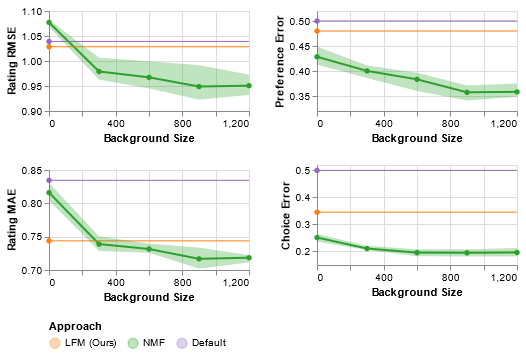}
    \caption{
        Performance (RMSE, MAE, and error rate) for all tasks with LFM (using Llama 2 13B) and NMF using different background sizes with history size 30.
    }
    \label{fig:alltask_30history_permodel_allbg_errorreliability}
\end{figure}

In Figure \ref{fig:alltask_30history_permodel_allbg_errorreliability} we compare LFM against NMF with increasing amounts of background data.
With background data we refer to rating histories from other users which can be used to improve the NMF embedding model.

The graph shows that using this background data greatly improves the performance of NMF, while LFM and Direct-LLM are zero-shot models that do not take advantage of this data.
We see this inability to take advantage of background data as one of the biggest shortcomings of zero-shot LFM, since it can neither tune the decoder for the optimal prediction (e.g., predict expectation of rating value instead of integer ratings), nor can it learn properties of the current corpus (e.g., more critical users that give lower ratings).



More detailed results are available in Appendix \ref{appendix:experimentsfull}.

\section{Discussion and Directions}

Regarding our main research question, we find that generating text-based profiles in LFM is competitive in accuracy with LLM approaches that do direct prediction across a range of tasks.
Furthermore, at least in principle LFM can reduce the latency of predictions depending on profile length and user history size, since profiles can be computed offline and the decoder LLM can use a substantially smaller prompt length than the Direct LLM.
This provides evidence that the new affordances (e.g., interpretability, steerability) of text-based profiles may not come at a big expense compared to other LLM approaches for recommendation.

However, we conclude that our current zero-shot approach to using LLMs for recommendation needs to be improved through task-directed training.
Zero-shot LLMs do not always provide readable outputs and definitely not brief outputs, they are biased due to a lack of adaptation to the task at hand, and they cannot make use of increasing amounts of background data.
We thus plan to investigate approaches for fine-tuning the LFM pipeline on task-specific data and formats.
We also aim to incorporate other textual metadata for both items and users, such as outside comments, tags, or transcriptions of the media itself.

Finally, we are keen to explore usability issues in how the textual profiles can be interpreted by people and how helpful they are for accessibility and steerability (e.g. manual interventions).

\begin{acks} 
This research was supported in part by the Graduate Fellowships for STEM Diversity (GFSD), as well as NSF Awards IIS-1901168, IIS-2008139 and IIS-2312865.
All content represents the opinion of the authors, which is not necessarily shared or endorsed by their respective employers and/or sponsors.
\end{acks}

\bibliographystyle{ACM-Reference-Format}
\bibliography{ref}

\appendix

\section{Runtime}
\label{appendix:runtime}

We logged the approximate runtime of different experiment stages for Llama 2 with entirely sequential (unoptimized) inference on one third of our task dataset.
Actual runtimes may vary heavily depending on optimization practice and other environment variables.

\begin{table}
    \caption{
        Sample approximated runtime of different pipeline stages on an RTX A6000, sequential (unbatched and unoptimized) prediction using Llama 2 13B for all language model approaches. History size 10.
    }
    \label{tab:methods_runtime}
    \begin{tabular}{c || c | c | c | c}
        \toprule
        Approach & Summarizing & Rating  & Pref.   & Choice  \\
                & (n=300)      & (n=900) & (n=900) & (n=900) \\
        \midrule
        LFM 50     & 25 min & 55 min & 80 min  & 80 min  \\
        LFM 100    & 25 min & 60 min & 100 min & 100 min \\
        LFM 200    & 35 min & 60 min & 80 min  & 100 min \\
        Direct     & -      & 50 min & 120 min & 100 min \\
        NMF bg1200 & 10 sec & 6 sec  & 6 sec   & -       \\
        (Ratings)  &        &        &         &         \\
        NMF bg1200 & 90 sec & -      & -       & 10 sec  \\
        (Choices)  &        &        &         &         \\
        \bottomrule
    \end{tabular}
\end{table}

\section{Task Prediction Extraction}

The output of a language model is in plain text format and contains more tokens than the prediction alone, so we need to extract the prediction values from the model output.

For the rating prediction task, we implement a naive regular expression function to identify cases where the predicted rating score is clear.
This regex captures a range of patterns, including but not limited to ``\texttt{[score]/5}'', ``\texttt{[score] out of 5}'', ``\texttt{a rating of [score]}''.

It then checks if the score found in all of these patterns are matched, which avoids producing wrong extraction given outputs such as ``I would give this movie a score of 3 or 4 out of 5''.

If the score is consistent, it finally checks if the score is within reasonable bounds and returns the score.

For the pairwise preference prediction task, we observed the complexity in the model outputs when describing its preferences. Thus, we passed the first model outputs into another LLM call to extract the exact preference. Then, we used a similar regex approach to process the final outputs.

For the pairwise choice task, we again implemented a regular expression function directly on output text.

We acknowledge that are definitely weaknesses in this approach, primarily in the possibility for these prediction extraction methods to have disparate performance across different approaches (e.g. it may parse LFM output text worse than LLM-Direct output text).
This is arguably a weakness that is tied to the challenges of using zero-shot models in general for these tasks: there is no consistent output format inherently learned, understood, and used by these models.

\section{Prompts and Hyperparameters}
\label{appendix:prompt}

The user preference learning prompt, as shown in Figure \ref{fig:prompt_pref_learn}, lists the user rating history and request the model to summarize the underpinning user preference.
The prediction prompt, in Figure \ref{fig:prompt_pref_pred}, uses the summarized user preference, and ask the rating for a new movie.
For the baseline task, in Figure \ref{fig:prompt_direct_pred}, we directly bypass the user preference learning and use history ratings to predict a new movie rating.

\begin{figure}
\begin{tcolorbox}[colback=white]
\begin{lstlisting}[breaklines, breakindent=0pt, basicstyle=\ttfamily\footnotesize]
[INST] I gave As Good as It Gets (1997) a rating of 3.0 out of 5.

I gave Fifth Element, The (1997) a rating of 3.0 out of 5.

I gave Ferris Bueller's Day Off (1986) a rating of 5.0 out of 5.

...

Based on this rating history, summarize the reasons why I like or dislike certain movies in under 100 words. Do not quote movie titles. [/INST]
\end{lstlisting}
\end{tcolorbox}
\caption{The prompt used for user preference summarization with length and content constraints.}
\label{fig:prompt_pref_learn}
\end{figure}

\begin{figure}
\begin{tcolorbox}[colback=white]
\begin{lstlisting}[breaklines, breakindent=0pt, basicstyle=\ttfamily\footnotesize]
[INST] You seem to enjoy movies with complex characters and engaging storylines, such as Ferris Bueller's Day Off and The Sixth Sense. You also prefer films with a mix of drama and comedy, like Good, the Bad and the Ugly and Mr. Holland's Opus. On the other hand, you are less fond of movies with overly complex plots or excessive violence, like The Abyss and Star Trek: First Contact. What score out of 5 would you give Die Hard (1988)? [/INST]
\end{lstlisting}
\end{tcolorbox}
\caption{The prompt used from user preference summarization to predict a new movie rating.}
\label{fig:prompt_pref_pred}
\end{figure}

\begin{figure}
\begin{tcolorbox}[colback=white]
\begin{lstlisting}[breaklines, breakindent=0pt, basicstyle=\ttfamily\footnotesize]
[INST] I gave As Good as It Gets (1997) a rating of 3.0 out of 5.

...

What score out of 5 would you give Die Hard (1988)? [/INST]
\end{lstlisting}
\end{tcolorbox}
\caption{The prompt used to directly predict a new movie rating using the user's rating history.}
\label{fig:prompt_direct_pred}
\end{figure}

\begin{figure}
\begin{tcolorbox}[colback=white]
\begin{lstlisting}[breaklines, breakindent=0pt, basicstyle=\ttfamily\footnotesize]
[INST] You seem to enjoy movies with complex characters and engaging storylines... Based on this user preference summary, guess which movie does the user prefer, A: Indiana Jones and the Last Crusade (1989) or B: In the Line of Fire (1993). Answer with 'A' or 'B'. [/INST]
\end{lstlisting}
\end{tcolorbox}
\caption{The prompt used from user preference summarization to predict a pairwise preference.}
\label{fig:prompt_preference_pred}
\end{figure}

\begin{figure}
\begin{tcolorbox}[colback=white]
\begin{lstlisting}[breaklines, breakindent=0pt, basicstyle=\ttfamily\footnotesize]
[INST] I gave As Good as It Gets (1997) a rating of 3.0 out of 5.

...

Based on this user rating history, guess which movie does the user prefer, A: Indiana Jones and the Last Crusade (1989) or B: In the Line of Fire (1993). Answer with 'A' or 'B'. [/INST]
\end{lstlisting}
\end{tcolorbox}
\caption{The prompt used to directly predict a pairwise preference using the user's rating history.}
\label{fig:prompt_direct_preference_pred}
\end{figure}

\begin{figure}
\begin{tcolorbox}[colback=white]
\begin{lstlisting}[breaklines, breakindent=0pt, basicstyle=\ttfamily\footnotesize]
[INST] You seem to enjoy movies with complex characters and engaging storylines... Based on the above user preference summary, guess which movie the user is more likely to have also consumed and reviewed, A: Indiana Jones and the Last Crusade (1989) or B: In the Line of Fire (1993). Answer with 'A' or 'B'. [/INST]
\end{lstlisting}
\end{tcolorbox}
\caption{The prompt used from user preference summarization to predict a movie choice.}
\label{fig:prompt_choice_pred}
\end{figure}

\begin{figure}
\begin{tcolorbox}[colback=white]
\begin{lstlisting}[breaklines, breakindent=0pt, basicstyle=\ttfamily\footnotesize]
[INST] I gave As Good as It Gets (1997) a rating of 3.0 out of 5.

...

Based on the above user rating history, guess which movie the user is more likely to have also consumed and reviewed, A: Indiana Jones and the Last Crusade (1989) or B: In the Line of Fire (1993). Answer with 'A' or 'B'. [/INST]
\end{lstlisting}
\end{tcolorbox}
\caption{The prompt used to directly predict a movie choice using the user's rating history.}
\label{fig:prompt_direct_choice_pred}
\end{figure}

\begin{figure}
\begin{tcolorbox}[colback=white]
\begin{lstlisting}[breaklines, breakindent=0pt, basicstyle=\ttfamily\footnotesize]
Based on your rating history, you tend to enjoy movies with complex plots and intricate storytelling, such as Big Lebowski, Burn After Reading, and Matrix. You also appreciate well-crafted action scenes and suspenseful moments, as seen in Neon Genesis Evangelion and The Punisher. However, you tend to dislike movies with overly simplistic or predictable plots, as well as those with excessive violence or gore, such as Surrogates and Once Upon a Time in Mexico.
\end{lstlisting}
\end{tcolorbox}
\caption{Example user summary (input size 10): demonstrates specific movie titles and non-genre movie attributes.}
\label{fig:summary_example_10}
\end{figure}

\begin{figure}
\begin{tcolorbox}[colback=white]
\begin{lstlisting}[breaklines, breakindent=0pt, basicstyle=\ttfamily\footnotesize]
You have a clear preference for well-reviewed, popular movies, with a strong rating for Interstellar (5/5) and a lower rating for Austin Powers: International Man of Mystery (3.5/5). You also enjoy action-packed movies, with a high rating for Kill Bill: Vol. 1 (5/5) and a lower rating for Balls of Fury (3.5/5). You have a soft spot for animated movies, with high ratings for Finding Nemo (5/5) and Ponyo (5/5). Additionally, you tend to enjoy psychological dramas, with high ratings for Beautiful Mind, A (4.5/5) and Prisoners (4.5/5).
\end{lstlisting}
\end{tcolorbox}
\caption{Example user summary (input size 20): demonstrates specific movie titles, specific ratings, and some broader genre descriptions.}
\label{fig:summary_example_20}
\end{figure}

\begin{figure}
\begin{tcolorbox}[colback=white]
\begin{lstlisting}[breaklines, breakindent=0pt, basicstyle=\ttfamily\footnotesize]
You enjoy movies with high ratings, such as The Imitation Game, Princess Bride, The Matrix, and Lord of the Rings. You dislike movies with low ratings, such as Star Wars: Episode II - Attack of the Clones, Pirates of the Caribbean: At World's End, and Suicide Squad. You also have mixed opinions on movies like Inglourious Basterds, The Incredibles, and Wonder Woman.
\end{lstlisting}
\end{tcolorbox}
\caption{Example user summary (input size 30): demonstrates specific movie titles, and seemingly nothing else.}
\label{fig:summary_example_30}
\end{figure}

All LLM model inferences were run with ``temperature'' 0.6, ``top\_p'' 0.9, ``top\_k'' 50, ``repetition\_penalty'' 1.2, empty system prompt, and otherwise default hyperparameters.
Some inferences were run with floating point type ``torch.float16'' and others were run with ``bfloat16'' due to occasional precision and GPU compatibility issues.

We tuned NMF hyperparameters by evaluating accuracy on rating prediction in a validation dataset. In all of our final experiments, we use unbiased NMF with 15 factors and 10 epochs.

\section{Example Summaries} \label{app:summaries}

It would be remiss to build a system focused on improving transparency by generating intermediate human-readable summaries without at least briefly examining the summaries themselves.

We sampled the user summary content and examined what types of information they contain.
Across all summaries for all input sizes, it is common to see direct movie titles mentioned (despite the prompt instructing Llama 2 to avoid using movie titles), as well as various movie attribute or tag type descriptions.
Some summaries included mentions of what exact score the user gave to each quoted movie.
In the future, we would like to do a more rigorous analysis of how this content may vary across different input sizes.



\begin{figure}
    \centering
    \includegraphics[width=\linewidth]{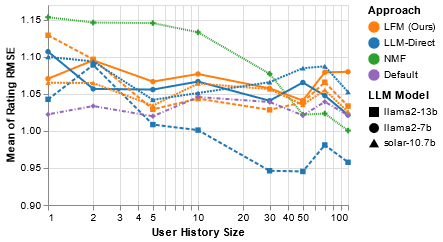}
    \caption{
    RMSE of rating prediction across all approaches and models, with varying user history size.
    }
    \label{fig:bonus_score_rmse}
\end{figure}

\begin{figure}
    \centering
    \includegraphics[width=\linewidth]{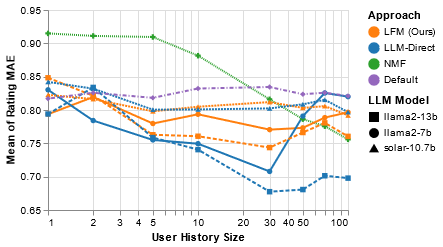}
    \caption{
    MAE of rating prediction across all approaches and models, with varying user history size.
    }
    \label{fig:bonus_score_mae}
\end{figure}

\begin{figure}
    \centering
    \includegraphics[width=\linewidth]{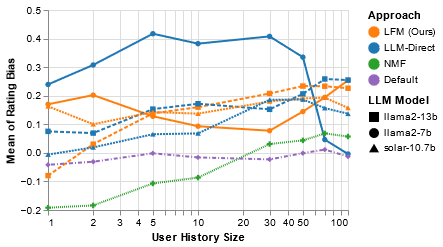}
    \caption{
    Bias of rating prediction across all approaches and models, with varying user history size.
    }
    \label{fig:bonus_score_bias}
\end{figure}

\section{More Experiment Results}
\label{appendix:experimentsfull}

\begin{figure}
    \centering
    \includegraphics[width=\linewidth]{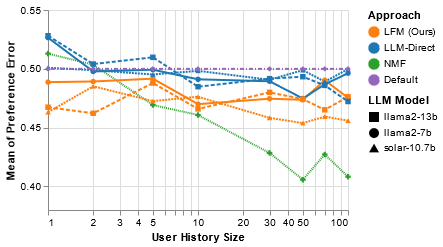}
    \caption{
    Error of preference prediction across all approaches and models, with varying user history size.
    }
    \label{fig:bonus_pref_error}
\end{figure}

\begin{figure}
    \centering
    \includegraphics[width=\linewidth]{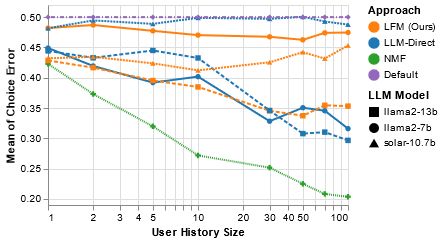}
    \caption{
    Error of choice prediction across all approaches and models, with varying user history size.
    }
    \label{fig:bonus_choice_error}
\end{figure}

Figures \ref{fig:bonus_score_rmse}, \ref{fig:bonus_score_mae}, \ref{fig:bonus_pref_error}, \ref{fig:bonus_choice_error}, \ref{fig:bonus_score_bias} show more detailed experimental results for all tasks.

For rating prediction, LLMs tend to have comparable performance when using LFM vs. direct prediction.
All language models still tend to show more positive bias compared to NMF.
In particular, Llama 2 7b using direct prediction demonstrates very high bias when its output texts are reliable, and then plummets to low bias when all output texts have become unreliable and all inferences are replaced by imputed guesses instead.

For preference prediction, LFM tends to perform better than LLM-direct across all models, and demonstrates the same cold-start benefit compared to NMF.

For choice prediction, LFM tends to have comparable or slightly worse error rate compared to LLM-direct.

\end{document}